\documentclass[11pt]{article}
\usepackage[utf8]{inputenc}
\usepackage{microtype}
\usepackage{amsmath}
\usepackage{amssymb}
\usepackage{mathtools}
\usepackage{graphicx}
\usepackage{algorithm}
\usepackage{algorithmic}
\usepackage{tikz}
\usepackage{pgfplots}
\usepackage{booktabs}
\usepackage{subcaption}
\usepackage{multirow}
\usepackage{xcolor}
\usepackage{listings}
\usepackage[round]{natbib}
\usepackage{url}

\usepackage[colorlinks=true,citecolor=blue,linkcolor=blue,urlcolor=blue]{hyperref}
\usepackage[capitalize,noabbrev]{cleveref}

\newcommand{\single}{\mathbf{s}}
\newcommand{\pair}{\mathbf{p}}
\newcommand{\density}{\mathbf{D}}
\newcommand{\MoleculePairformer}{\textsc{MoleculePairformer}}

\newcommand{\Normal}{\mathcal{N}}

\newcommand{\Angstrom}{\text{\AA}}

\DeclareMathOperator*{\argmin}{arg\,min}
\DeclareMathOperator{\softmax}{softmax}

\usepackage[top=1in, bottom=1in, left=1.25in, right=1.25in]{geometry}

\usepackage{pdflscape}

\usepackage{eso-pic}
\newcommand{\lscapepagenumber}{%
  \AddToShipoutPictureFG*{%
    \AtPageLowerLeft{%
      \put(\LenToUnit{\dimexpr\paperwidth-0.7cm\relax},\LenToUnit{0.5\paperheight}){%
        \makebox(0,0){\rotatebox{90}{\thepage}}}}}%
}

\title{\Large\bf Sesame: Structure-Aware Molecular Generation via Spatial Density-Map Conditioning}

\author{
Konstantin Yatsenko, Arvind Thiagarajan\thanks{Corresponding author: \texttt{arvind@tessel.bio}} \\
{\footnotesize Tessel Biosciences, Inc., 750 Main Street, Cambridge, MA 02139} \\
}

\date{}

\begin{document}

\maketitle
\thispagestyle{empty}

\begin{abstract}
\noindent
Generative molecular models for drug design are a promising direction with much active research.
In the next phase of computational drug design, such models will need to understand small molecule structure and protein-ligand interactions, and they will need to possess the machinery
to generate molecules \textit{de novo}. Incorporating each feature poses a critical challenge. 
Equally important, yet often treated as secondary, is the ability to grow a molecule from a partial starting
point -- a scaffold or fragment supplied by a chemist -- which is the central operation of lead optimization.
We present Sesame (Spatial Evoformer for a Structure-Aware Molecular Engine), a diffusion-based molecular generation model that leverages a novel spatial pairformer module
to condition on partial molecular structure and the surrounding protein pocket, both expressed as continuous spatial
density maps. This single conditioning mechanism supports both \textit{de novo} generation and fragment-conditioned
lead optimization, letting a medicinal chemist prune a hit to a scaffold and have Sesame grow it in productive ways.
In addition to this module, we also introduce a diffusion
framework for joint denoising of atom types, bond types, and positions, along with a trajectory finetuning scheme that
trains on the model's own sampling rollouts to improve generation quality. Sesame is trained on a large corpus of ligand-only and
protein-ligand datasets.
\end{abstract}


\begin{figure}[htbp]
\centering
\includegraphics[width=0.95\textwidth]{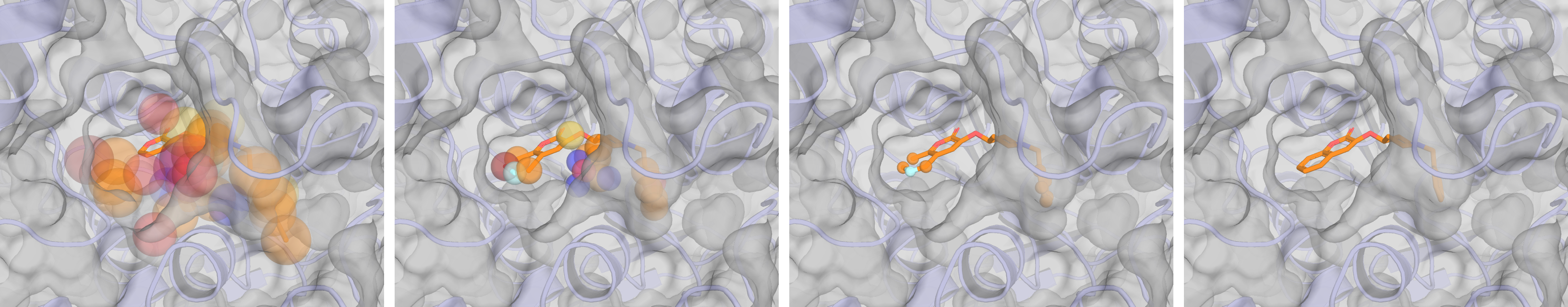}
\caption{Snapshot of the diffusion process, left-to-right. Here, we denoise a ligand with fragment conditioning, allowing the model to fill in the gaps and extend the existing core.}
\label{fig:diffusion_series}
\end{figure}

\section{Introduction}
\label{sec:introduction}

\subsection{Problem Statement}
Modern protein-folding models possess capabilities to reason about protein-ligand interactions in a meaningful way, with
good accuracy across a number of tasks. However, these models are fundamentally constrained by knowledge about a specific
ligand, often passed as hard constraints for the generation algorithm. These architectures are incompatible with the
generation of \textit{de novo} molecules, which means that any system using these models for drug discovery must carefully
balance exploration of chemical space with computational costs and constraints. Furthermore, these
protein-ligand models are not perfect, and can often generate structures with small inconsistencies, e.g.\ incorrect enantiomers.
Our approach aims to learn from successes in the protein-folding space, extending them to perform full
molecular generation.

Beyond \textit{de novo} generation, a second capability is equally central to real drug-discovery campaigns and is
often underserved by generative models: lead optimization, in which a known hit is refined rather than designed from
scratch. In practice this is where human expertise is most valuable -- a medicinal chemist can identify the substructure
of a hit worth preserving and prune the molecule down to that scaffold. A useful generative model should treat such a
fragment as a soft prior to be grown and elaborated, not as a rigid atom-level constraint, so that human insight and
generative chemistry compose rather than compete. We therefore treat fragment-conditioned generation as a primary
objective alongside \textit{de novo} design, and, as described below, realize both through a single conditioning mechanism.

\subsection{Motivation}
Our approach is motivated by three key insights. First, encoding the local structure of a protein by estimating various
forces on a dense grid is a common technique, and largely used in various docking algorithms. These so-called density maps
provide a natural representation for 3D structural information, encoding multiple physical properties
(charge, hydrophobicity, hydrogen bonding, van der Waals interactions) in a unified grid-based
format. This allows for modeling diverse sets of protein pockets with identically-sized inputs, which confers a number of
computational efficiencies. Second, Pairformer architectures perform admirably for a variety of tasks that require complex
multi-way interactions between components. An extension to this architecture to allow for information to pass from the
density map to the prediction is an essential component to a model whose inputs are density maps. Third, diffusion models
whose goal is to produce molecules require a diffusion approach that handles both discrete atom types and continuous atomic
coordinates. Many naive approaches fail to take into account biases that trained models produce, and a careful examination
of the reverse diffusion process is necessary to have a robust generative model.

\subsection{Contributions}
Our main contributions are:
\begin{enumerate}
    \item \textbf{Novel density map conditioning architecture}: An attention-based mechanism that
    adaptively samples from density maps to guide molecular generation, enabling structure-aware generation.
    \item \textbf{Unified fragment and pocket conditioning for lead optimization}: Because both the protein pocket and a
    partial ligand are supplied to the model as the same continuous density field, a single mechanism supports
    \textit{de novo} generation and fragment-conditioned growth with no architectural changes. This makes lead
    optimization a first-class capability: a medicinal chemist can prune a hit molecule down to a scaffold, and Sesame
    grows chemically sensible, pocket-compatible completions around it -- a direct integration of human insight with
    generative chemistry. Across fragment-conditioned generation, 94.8\% of generated molecules retain the seeding
    fragment as a substructure, confirming that the conditioning is honored rather than ignored.
    \item \textbf{Hybrid discrete-continuous diffusion}: A unified diffusion process that
    handles discrete atom types and bond types via categorical diffusion and continuous coordinates via Gaussian diffusion.
    \item \textbf{Trajectory finetuning}: A self-distillation finetuning scheme that rolls out the model's own reverse-diffusion trajectories, re-matches the predicted atoms to ground truth, and trains on the resulting (noisy state, aligned target) pairs, improving the quality of fully generated molecules.
\end{enumerate}

\section{Related Work}
\label{sec:related}

\paragraph{Generating 3D molecules.}
Denoising generative models operating directly on 3D molecular structure have become the dominant paradigm for de novo design. Equivariant diffusion models established the template by jointly denoising atomic coordinates and types under E(3)/SE(3) symmetry \citep{hoogeboom2022edm}, with subsequent work unifying the diffusion of continuous coordinates and discrete atom and bond features \citep{vignac2023midi}. A complementary line replaces denoising diffusion with flow matching, trading stochastic reverse-time dynamics for deterministic transport and typically improving sampling efficiency and physical quality \citep{irwin2025semlaflow, dunn2026flowmol3}. Most of these methods target generic small molecules -- frequently benchmarked on QM9-scale data -- rather than drug-like chemistry per se \citep{hoogeboom2022edm, vignac2023midi, irwin2025semlaflow}, and they share a structural limitation relevant to our setting: the number of atoms is fixed prior to sampling, so the model cannot grow or trim a molecule to fit a target. A distinct architectural lineage comes not from molecule generation but from biomolecular structure prediction: AlphaFold3 couples a Pairformer trunk to a diffusion module that denoises raw atomic coordinates \citep{abramson2024af3}, and the open Boltz models adopt and extend this co-folding architecture \citep{wohlwend2024boltz1, passaro2025boltz2}. While these systems predict the structure of a given complex rather than generating novel ligands, their pairwise-representation-plus-diffusion-decoder design directly informs the architecture of Sesame.

\paragraph{Spatial-field and voxel representations.}
A smaller body of work represents molecules not as point sets but as continuous
spatial maps -- most directly, voxel grids denoised in a learned density space,
for both unconditional 3D generation \citep{pinheiro2024voxel} and
structure-based design \citep{pinheiro2024sbdd}. This framing is the closest
precedent for our conditioning scheme, in which both the pocket and the
(partial) ligand are expressed as maps of physical potentials. ShEPhERD
similarly operates over interaction fields -- shape, electrostatic potential, and
pharmacophores -- generating molecules matched to a target profile
\citep{adams2025shepherd}, a representation that is conceptually the inverse of
our pocket-conditioned setup.

\paragraph{Conditioning on scaffolds and fragments.}
Several methods generate molecules from a fixed structural prior -- decorating a
scaffold, linking fragments, or growing from a seed. DiffLinker designs linkers
conditioned on disconnected 3D fragments \citep{igashov2024difflinker}; DiffDec
performs structure-aware scaffold decoration end-to-end \citep{xie2024diffdec};
and 3D-Scaffold generates 3D coordinates of drug-like molecules around a
specified scaffold \citep{joshi2021scaffold}. These correspond to the second
conditioning mode in Sesame, in which a full or partial map of a scaffold or
fragment seeds generation; unlike methods that fix the substructure as discrete
atoms, Sesame supplies it as the same continuous field used for the pocket, allowing
the prior to be partial and the surrounding chemistry to be grown to fit.

\paragraph{Generative models for drug discovery.}
The methods above are largely concerned with generation in broad chemical space;
a parallel body of work shares our specific goal of discovering drug-like hits
against a target. For target-aware generation, diffusion models conditioned on
the protein binding site generate ligands directly within the pocket: TargetDiff
jointly models ligand atoms and their geometry conditioned on the protein
context and additionally estimates binding affinity \citep{guan2023targetdiff},
while DiffSBDD applies equivariant diffusion to structure-based design across
pocket-conditioned generation and inpainting tasks \citep{schneuing2024diffsbdd}.
More recent systems extend pocket conditioning to flow matching with
multi-objective guidance \citep{cremer2024pilot} and to unified interaction- and
fragment-based generation \citep{cremer2026flowr}. Where an explicit pocket is
unavailable, ShEPhERD pursues the same target-driven goal by conditioning on the
interaction profile of a known ligand and generating bioisosteric replacements
\citep{adams2025shepherd}. A second, persistent gap between these generative
models and practical hit finding is synthesizability: structurally valid
molecules are not necessarily makeable. A growing line of work addresses this by
generating directly in synthesizable chemical space -- composing molecules from
purchasable building blocks and reaction templates rather than from atoms
\citep{gao2025synthesizable}, or by amortizing the search over synthetic routes
with generative flow networks \citep{cretu2025synflownet}. Sesame differs from
the pocket-conditioned methods in that the pocket enters as a spatial potential
map rather than as an explicit atomistic context, and pocket conditioning is
optional rather than required; synthesizability-aware generation we view as
complementary to spatial-field conditioning rather than competing with it,
composable as a downstream filter or alternative decoder. Across these lines of
work, conditioning on a spatial field, growing from a partial fragment, and
targeting a protein pocket have been pursued largely in isolation; Sesame is
designed to support all three within a single generative process.

\section{Methods}
\label{sec:methods}

\subsection{Overall Architecture}
\label{subsec:overall_arch}

Our training pipeline consists of several key components: data generation from multiple sources,
density map computation, model forward pass, diffusion process, and loss computation.

The training process begins with data generation from two sources: ZINC (ligand-only) and SAIR (protein-ligand pairs).
Each sample undergoes preprocessing to generate density maps and extract molecular graphs. During training, the model
receives noisy atom types and positions from the forward diffusion process, along with density maps and conditioning
signals. The model predicts denoised atom types, positions, and bonds, which are compared to ground truth through a
multi-component loss function.

\subsection{MoleculePairformer Architecture}
\label{subsec:architecture}

\begin{table}[htbp]
\centering
\caption{Model dimensions and hyperparameters}
\label{tab:model_dims}
\begin{tabular}{lr}
\toprule
Parameter & Value \\
\midrule
Single dimension ($d_s$) & 384 \\
Pair dimension ($d_p$) & 128 \\
Time dimension ($d_t$) & 256 \\
Density channel dimension ($d_d$) & 384 \\
Number of layers ($L$) & 24 \\
Number of attention heads & 4 \\
Number of sampling points ($O \cdot H$) & 1024 \\
Atom types ($AT$) & 17 \\
Bond types ($BT$) & 6 \quad (types 1--5 used; see text) \\
Batch size ($B$) & 128 \\
Max atoms per molecule ($N$) & 50 \\
Density grid resolution & 0.5 \Angstrom \\
Density grid cube side length & 16 \Angstrom \\
Optimizer & AdamW \\
Weight decay & 0.1 \\
Learning rate & $2 \times 10^{-3}$ \\
\bottomrule
\end{tabular}
\end{table}

\subsubsection{Core Components}
\label{subsubsec:core_components}

The \MoleculePairformer{}\footnote{The \MoleculePairformer{} follows the Evoformer/Pairformer lineage of AlphaFold; architecturally it is a pairformer -- operating on single and pair representations with no MSA track -- and we refer to it as such throughout. The ``Evoformer'' in Sesame's expansion (Spatial Evoformer for a Structure-Aware Molecular Engine) reflects this lineage.} operates on two main representations (Figure~\ref{fig:pairformer_layer}A):
\begin{itemize}
    \item \textbf{Single representation} $\single \in \mathbb{R}^{N \times d_s}$: Per-atom features of dimension
    $d_s = 384$, where $N$ is the total potential number of atoms.
    \item \textbf{Pair representation} $\pair \in \mathbb{R}^{N \times N \times d_p}$: Atom-pair features of
    dimension $d_p = 128$.
\end{itemize}

Atom types are represented as 16 discrete classes corresponding to (element, number of implicit hydrogens) pairs
that occur in drug-like molecules, plus one additional pseudo-None class, giving $AT = 17$ total atom types.
The 16 drug-like classes are:
(C,0), (C,1), (C,2), (C,3), (N,0), (N,1), (N,2), (O,0), (O,1),
(F,0), (P,0), (S,0), (S,1), (Cl,0), (Br,0), (I,0).
These cover the vast majority of heavy atoms in drug-like molecules, with implicit hydrogen count
encoding local valence context without requiring explicit hydrogen atoms in the model.
One key feature of the molecular generation process is that the model needs to predict not only which atom types
exist where in space, but also how many atoms are needed for a given molecule. Put differently, the model may output
a maximum of $N$ atoms, but must predict the pseudo-None type for all atoms that do not exist in the final molecule.

The model uses multiple embedding layers:
\begin{enumerate}
    \item \textbf{Atom type embedding}: Linear layer mapping from $AT$ noised atom types to $d_s$ dimensions.
    \item \textbf{Atom position embedding}: Fourier embedding of atomic positions, followed by a linear
    projection to $d_s$ dimensions.
    \item \textbf{Bond type embedding}: Linear layer mapping from $BT$ noised bond types to $d_p$ dimensions,
    added to the position pair embedding.
    \item \textbf{Position pair embedding}: Fourier embedding of relative atomic positions, followed by a
    linear projection to $d_p$ dimensions. Diagonal elements receive the atom position instead of the (zero) relative position.
    \item \textbf{Time embeddings}: Three separate Fourier embeddings of the diffusion timestep $\tau \in [0,1]^3$,
    one per noise channel (coordinate, atom type, bond type), each projected to $d_t$ dimensions.
\end{enumerate}

The Fourier embedding for a scalar $t$ is defined as:
\begin{equation}
\text{Fourier}(t) = \sqrt{2} \cos(2\pi (t \mathbf{w} + \mathbf{b}))
\end{equation}
where $\mathbf{w}$ and $\mathbf{b}$ are learnable parameters.

Notably, for both the position and position pair embeddings, all 3 spatial dimensions are embedded with the same
$\mathbf{w}$ and $\mathbf{b}$. The linear projection that follows does distinguish between the 3 dimensions, though, so
the symmetry between dimensions ends there.

The atom type and atom position embeddings are summed to form the initial single representation $\single$.
The position pair embedding and bond type embedding are summed to form the initial pair representation $\pair$.
The coordinate and atom type time embeddings are each incorporated into the single representation via separate \textit{GatedResidual} layers, while the bond type time embedding is incorporated into the pair representation via a \textit{GatedResidual} layer.

Finally, each initial $6 \times 32 \times 32 \times 32$ density map (protein and ligand) is processed into a
$d_d \times 12 \times 12 \times 12$ volume via a series of 3D convolutions and MaxPool layers.

\subsubsection{Pairformer Layers}
\label{subsubsec:pairformer_layers}

Each \MoleculePairformer{} layer consists of several components applied in sequence:

\textbf{Density Map Conditioning Operations} (Figure~\ref{fig:pairformer_layer}C): First, we generate sampling points via attention on the single representation. Rather than self-attention, we use a per-layer learned key vector of shape $H\times O\times d_h$, where $O$ is the number of sampling points emitted per attention head. These then attend to the query vectors derived from the single representation. The value vectors (of shape $H \times 3$, where $H$ is the number of heads, typically 4) are also derived from the single representation. With $O = 256$ sampling points per head and $H = 4$ heads, this yields $1024$ sampling points in total.

The sampling points are used to extract features from the density map via trilinear interpolation:
\begin{equation}
\mathbf{F}_{\text{density}} = \text{GridSample}(\density_{\text{protein}}, \mathbf{P}_{\text{sample}})
\end{equation}
where $\density_{\text{protein}} \in \mathbb{R}^{d_d \times 12 \times 12 \times 12}$ is the post-3D convnet density map.

The $d_d$ density features are concatenated with sampling point coordinates
and used in cross-attention to update single and pair representations.

Queries are single- and pair-derived vectors, while key and value vectors are derived from the sampled points. Key vectors are shared between both representations, while value vectors are computed separately.

Finally, these attention terms are added back to the model via gated residual connections. Details of this module can be found in Appendix~\ref{app:density_map}.

\textbf{Triangle Operations}: Following AlphaFold's Evoformer architecture, we use triangle multiplication and triangle
attention operations. Triangle multiplication computes:
\begin{equation}
\mathbf{z}_{ij} = \sum_k \mathbf{a}_{ik} \mathbf{b}_{jk}
\end{equation}
for outgoing triangles, and similarly for incoming triangles.

\textbf{Triangle Attention}: Applies attention along one dimension of the pair representation:
\begin{equation}
\text{Attention}(\mathbf{Q}, \mathbf{K}, \mathbf{V}) = \softmax\left(\frac{\mathbf{Q}\mathbf{K}^T}{\sqrt{d_k}} + \mathbf{B}\right)\mathbf{V}
\end{equation}
where $\mathbf{B}$ is a pair bias learned from the pair representation.

\textbf{Attention with Pair Bias}: The single representation is updated via attention,
with bias terms from the pair representation:
\begin{equation}
\single' = \single + \text{Attention}(\single, \single, \single; \text{bias}=\pair)
\end{equation}

\textbf{Feed-Forward Networks}: We use SwiGLU~\citep{shazeer2020glu} activation:
\begin{equation}
\text{SwiGLU}(\mathbf{x}) = \text{SiLU}(\mathbf{W}_1 \mathbf{x}) \odot (\mathbf{W}_2 \mathbf{x})
\end{equation}

All operations are wrapped with residual connections and pre-layer normalization. Figure~\ref{fig:pairformer_layer}B
illustrates the layer structure.

\begin{landscape}
\thispagestyle{empty}
\lscapepagenumber
\begin{figure}
\centering
\includegraphics[width=1.5\textheight]{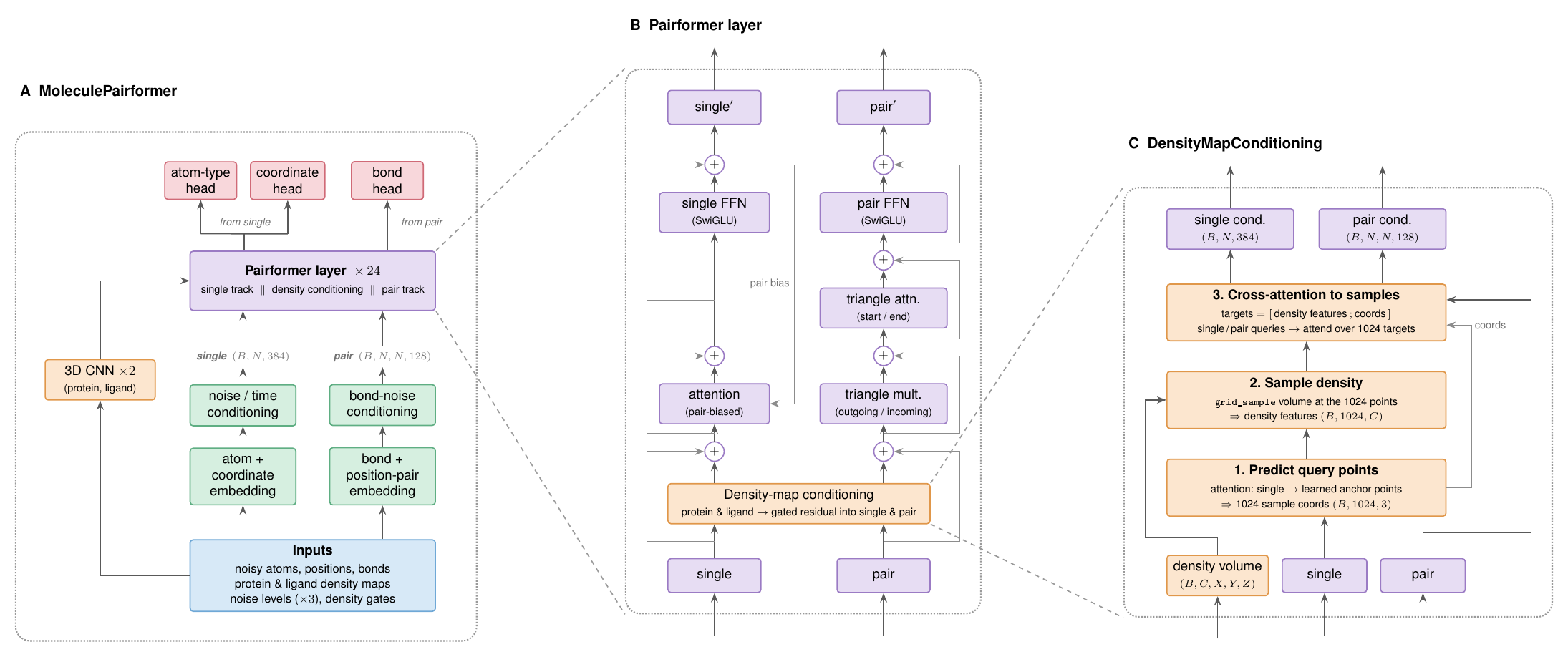}
\caption{The \MoleculePairformer{} architecture. \textbf{(A)} Noisy atoms, positions,
and bonds are embedded into single and pair representations, conditioned on noise level
and encoded protein/ligand density maps, and passed through $L=24$ Pairformer layers
before the atom-type, coordinate, and bond output heads. \textbf{(B)} A single Pairformer
layer applies density-map conditioning, then triangle and feed-forward updates on the pair
track and pair-biased attention on the single track. \textbf{(C)} Density-map conditioning:
the single representation predicts query points that sample the density volume, whose
features then update both tracks via cross-attention.}
\label{fig:pairformer_layer}
\end{figure}
\end{landscape}

\subsubsection{Output Heads}
\label{subsubsec:output_heads}

The model produces three outputs:

\textbf{Atom Type Head}: Predicts logits over $AT$ atom types:
\begin{equation}
\mathbf{A}_{\text{pred}} = \text{MLP}(\single)
\end{equation}

\textbf{Coordinate Head}: Predicts 3D coordinates:
\begin{equation}
\mathbf{C}_{\text{pred}} = \text{MLP}(\single)
\end{equation}

\textbf{Bond Head}: Predicts bond types between atom pairs:
\begin{equation}
\mathbf{B}_{\text{pred}} = \text{Symmetric}(\text{MLP}(\pair))
\end{equation}
where $\text{Symmetric}$ ensures the bond matrix is symmetric (bond between atoms $i$ and $j$ equals bond between $j$ and $i$).

\subsection{Diffusion Process}
\label{subsec:diffusion}

Molecular generation requires handling both discrete atom types and continuous atomic coordinates.
We introduce a novel diffusion process that jointly models these two modalities within a unified
framework.

The forward diffusion process adds noise independently to each modality, 
while the reverse process uses joint model predictions to denoise both simultaneously.

\subsubsection{Forward Process}
\label{subsubsec:forward}

The forward process defines a distribution $q(\mathbf{C}_\tau, \mathbf{A}_\tau, \mathbf{B}_\tau | \mathbf{C}_0, \mathbf{A}_0, \mathbf{B}_0, \tau)$
over noisy coordinates $\mathbf{C}_\tau$, noisy atom types $\mathbf{A}_\tau$, and noisy bond types $\mathbf{B}_\tau$,
given true atom coordinates $\mathbf{C}_0$,
true atom types $\mathbf{A}_0 \in \{0, \ldots, AT-1\}^N$, true bond types $\mathbf{B}_0 \in \{0, \ldots, BT-1\}^{N \times N}$,
at timestep $\tau \in [0, 1]^3$, where $\tau[0]$ controls coordinate noise, $\tau[1]$ controls atom type noise,
and $\tau[2]$ controls bond type noise, each sampled independently during training.
At $\tau = (1, 1, 1)$, all modalities are considered fully noised.

\paragraph{Coordinate diffusion}

For coordinates, we perform a relatively simple set of operations. First, we add noise sampled from
$\Normal\left(0, \sigma^2(\tau[0])\mathbf{I}\right),$ where $\sigma: [0, 1] \to [s_{\min}, s_{\max}]$ is a
power-law function, with $p = 7,$ $s_{\min} = 0.01 \Angstrom,$ and $s_{\max} = 8 \Angstrom$.
\begin{equation}
\sigma(\tau[0]) = \left(s_{\max}^{1/p} + (1-\tau[0])(s_{\min}^{1/p} - s_{\max}^{1/p})\right)^p
\end{equation}
Because coordinate noise is added independently to each atom, a freshly noised cloud is not arranged
the way the model expects to denoise it: the point that lands near a given ground-truth atom is not
necessarily the one sharing its index. We therefore re-pair the noised points to the ground-truth atoms
by minimum total displacement, using a modified Jonker-Volgenant assignment, so that the noisy state is
already in correspondence with the target the model predicts. Importantly, this re-pairing does not move
any points; as an unordered cloud the noised positions are unchanged, and the assignment only selects
which ground-truth atom each point denoises toward.

Concretely, let $\tilde{\mathbf{C}}_\tau = \mathbf{C}_0 + \boldsymbol{\epsilon}$ with
$\boldsymbol{\epsilon} \sim \Normal(0, \sigma^2(\tau[0])\mathbf{I})$ be the noised positions. These are
sampled independently, giving the product form
\begin{equation}
q(\tilde{\mathbf{C}}_\tau \mid \mathbf{C}_0, \tau)
= \prod_{i=1}^N \Normal\!\left(\tilde{\mathbf{C}}_\tau[i];\, \mathbf{C}_0[i],\, \sigma^2(\tau[0])\mathbf{I}\right).
\end{equation}
We then compute the pairwise distance cost matrix $\text{Cost}_{ij} = \|\tilde{\mathbf{C}}_\tau[i] - \mathbf{C}_0[j]\|_2$
and find the assignment minimizing total displacement:
\begin{equation}
\pi^* = \argmin_{\pi \in S_N} \sum_{i=1}^N \|\tilde{\mathbf{C}}_\tau[\pi(i)] - \mathbf{C}_0[i]\|_2,
\end{equation}
which pairs each ground-truth atom $i$ with the noised point $\pi^*(i)$ it should denoise toward.
The matched coordinates are $\mathbf{C}_\tau[i] = \tilde{\mathbf{C}}_\tau[\pi^*(i)]$, and the discrete
targets are permuted by the same assignment ($\mathbf{A}_0$ reindexed by $\pi^*$, and the rows and columns
of $\mathbf{B}_0$ permuted identically), so that type and bond identity remain consistent with the
re-paired positions. The discrete corruptions are then applied to these permuted targets:
\begin{align}
q(\mathbf{A}_\tau \mid \mathbf{A}_0, \pi^*, \tau) &= \prod_{i=1}^N q(a_\tau[i] \mid a_0[i], \tau[1]), \\
q(\mathbf{B}_\tau \mid \mathbf{B}_0, \pi^*, \tau) &= \prod_{i=1}^N \prod_{j=1}^N q(b_\tau[i,j] \mid b_0[i,j], \tau[2]),
\end{align}
where $a_0[i]$ and $b_0[i,j]$ denote the permuted ground-truth types
($\mathbf{A}_0[\pi^*(i)]$ and $\mathbf{B}_0[\pi^*(i), \pi^*(j)]$), and the per-atom and per-pair
discrete marginals are as defined in the paragraphs below.

\paragraph{Atom Type Diffusion}

We use a marginal diffusion scheme, following the framework of DiGress~\citep{vignac2022digress}.
The atom type noise level is parameterized by $\tau[1] \in [0, 1]$ via a cosine schedule:
\begin{equation}
\bar\alpha(\tau[1]) = \cos^2\!\left(\frac{\pi\,\tau[1]}{2}\right)
\end{equation}
so that $\bar\alpha(0) = 1$ (no noise) and $\bar\alpha(1) = 0$ (fully noised).
The forward marginal for a single atom type is then:
\begin{equation}
q(a_{\tau[1]} = k \mid a_0 = i) = \bar\alpha(\tau[1])\,\mathbf{1}[i=k] + (1 - \bar\alpha(\tau[1]))\,m_k^{(A)}
\end{equation}
where $m^{(A)}_k = 1/AT$ is the uniform distribution over atom types.
A uniform prior means that at full noise ($\bar\alpha = 0$) every atom type is equally likely,
avoiding any systematic bias toward chemically common types during the noising process.

For a pair of consecutive noise levels $\tau_i[1] > \tau_{i-1}[1]$, the effective one-step corruption probability is:
\begin{equation}
\beta_{\tau_{i-1} \to \tau_i} = 1 - \frac{\bar\alpha(\tau_i[1])}{\bar\alpha(\tau_{i-1}[1])}
\end{equation}
This ratio-based formulation means no fixed discrete schedule is required: during training, $\tau[1]$ is sampled
uniformly from $[0,1]$, and during generation a trajectory of sorted noise levels is used, with $\beta$ derived
from consecutive $\bar\alpha$ ratios at each step.

\paragraph{Bond Type Diffusion}

Bond types undergo the same marginal diffusion scheme as atom types, but use an independent noise
coordinate $\tau[2] \in [0,1]$ with its own cosine schedule $\bar\alpha(\tau[2]) = \cos^2(\pi\tau[2]/2)$.
Each pairwise bond type is independently noised as:
\begin{equation}
q(b_{\tau[2]} = k \mid b_0 = i) = \bar\alpha(\tau[2])\,\mathbf{1}[i=k] + (1 - \bar\alpha(\tau[2]))\,m_k^{(B)}
\end{equation}
where $m^{(B)}$ is a fixed prior over bond types, chosen to reflect the strong sparsity of molecular graphs:
\begin{equation}
m^{(B)} = [0,\; 0.0025,\; 0.0025,\; 0.0025,\; 0.0025,\; 0.99]
\end{equation}
corresponding to types 0 (unused), 1 (single), 2 (double), 3 (triple), 4 (aromatic), and 5 (None) respectively.
The dominant weight on type 5 reflects the fact that the vast majority of atom pairs in a molecule are unbonded.
The small equal weights on types 1--4 provide a weak non-zero prior over all bonded types, ensuring the
posterior never assigns exactly zero probability to any valid bond during the reverse process.
Type 0 receives zero prior weight, as it is unused.

The noised bond types $\mathbf{B}_\tau$ are provided as input to the model alongside the noised atom types $\mathbf{A}_\tau$,
encoded via the pair representation.

\subsubsection{Reverse Process}
\label{subsubsec:reverse}

The reverse process defines a distribution
$p(\mathbf{A}_{\tau_{i-1}}, \mathbf{B}_{\tau_{i-1}}, \mathbf{C}_{\tau_{i-1}} |\mathbf{A}_{\tau_i}, \mathbf{B}_{\tau_i}, \mathbf{C}_{\tau_i}, \theta, d)$ over denoised
atom type logits, bond type logits, and coordinates, given noisy inputs, model parameters $\theta$, and density maps $d$. The model predicts all three
jointly, but the denoising processes are conditionally independent given these predictions:

\begin{align}
&p(\mathbf{C}_{\tau_{i-1}}, \mathbf{A}_{\tau_{i-1}}, \mathbf{B}_{\tau_{i-1}} \mid \mathbf{A}_{\tau_i}, \mathbf{B}_{\tau_i}, \mathbf{C}_{\tau_i}, \theta, \mathbf{d}) \nonumber \\
&\quad = p(\mathbf{A}_{\tau_{i-1}} \mid \mathbf{A}_{\tau_i}, \mathbf{B}_{\tau_i}, \mathbf{C}_{\tau_i}, \theta, \mathbf{d}) \cdot
p(\mathbf{B}_{\tau_{i-1}} \mid \mathbf{A}_{\tau_i}, \mathbf{B}_{\tau_i}, \mathbf{C}_{\tau_i}, \theta, \mathbf{d}) \cdot
p(\mathbf{C}_{\tau_{i-1}} \mid \mathbf{A}_{\tau_i}, \mathbf{B}_{\tau_i}, \mathbf{C}_{\tau_i}, \theta, \mathbf{d})
\end{align}

\paragraph{Coordinates}

The model predicts denoised positions $\mathbf{C}_{\text{pred}} = g_\theta(\mathbf{A}_{\tau_i}, \mathbf{B}_{\tau_i}, \mathbf{C}_{\tau_i}, \tau_i, \mathbf{d})$.
Coordinates are updated deterministically, while the discrete atom and bond types are sampled from their reverse
posteriors (below); the coordinate update follows the DDIM-style probability-flow rule~\citep{song2021ddim}
for a variance-exploding schedule:

\begin{equation}
\mathbf{C}_{\tau_{i-1}} = \mathbf{C}_{\text{pred}} + \frac{\sigma(\tau_{i-1}[0])}{\sigma(\tau_i[0])} (\mathbf{C}_{\tau_i} - \mathbf{C}_{\text{pred}})
\end{equation}

Note that since the model is trained to predict positions post-matching, we assume that the model predictions already correspond
to the correct input atoms, and do not perform any matching algorithm during a denoising step.

\paragraph{Types}

The model predicts $x_0$ logits $\mathbf{A}_{\text{pred}} = h_\theta(\mathbf{A}_{\tau_i}, \mathbf{B}_{\tau_i}, \mathbf{C}_{\tau_i}, \tau_i, \mathbf{d})$
for atom types. The reverse posterior is derived by Bayes' rule and marginalizing over $x_0$:
\begin{equation}
p_\theta(a_{\tau_{i-1}} = j \mid a_{\tau_i} = k)
\;\propto\;
q(a_{\tau_i} = k \mid a_{\tau_{i-1}} = j)
\cdot \sum_{i'} \frac{p_\theta(a_0 = i' \mid a_{\tau_i})}{q(a_{\tau_i} = k \mid a_0 = i')} \cdot q(a_{\tau_{i-1}} = j \mid a_0 = i')
\end{equation}
where all three terms have closed forms under the marginal diffusion structure (Section~\ref{subsubsec:forward});
in particular, the one-step transition $q(a_{\tau_i} \mid a_{\tau_{i-1}})$ is the marginal-scheme transition with
corruption probability $\beta_{\tau_{i-1}\to\tau_i}$ defined there.
Critically, the $q(a_{\tau_i} = k \mid a_0 = i')$ denominator must remain inside the sum —
factoring it out drops an $i'$-dependent correction that systematically over-concentrates probability mass on the
most likely class, producing an $O(K)$ error on minority classes at low noise levels.

In practice this reduces to a $K \times K$ matrix multiply per atom:
let $\mathbf{w} = p_\theta(a_0 \mid a_{\tau_i}) / q(a_{\tau_i} \mid a_0)$ (element-wise, clamped for numerical stability),
and $[\mathbf{Q}_{\tau_{i-1}}]_{i'j} = q(a_{\tau_{i-1}} = j \mid a_0 = i')$; the unnormalized posterior is:
\begin{equation}
\tilde{p}_j \propto q(a_{\tau_i} = k \mid a_{\tau_{i-1}} = j) \cdot (\mathbf{w}^\top \mathbf{Q}_{\tau_{i-1}})_j
\end{equation}
normalized over $j \in \{0, \ldots, AT-1\}$, from which $a_{\tau_{i-1}}$ is sampled.

\paragraph{Bond Types}

Bond types are denoised using the identical posterior formula, with $m^{(B)}$, $BT$,
$\tau[2]$, and $\mathbf{B}_{\text{pred}}$ replacing their atom-type counterparts.

\subsection{Training Methodology}
\label{subsec:training}

\subsubsection{Diffusion Training Objective}
\label{subsubsec:diffusion_training}

During initial training, the model learns to predict both atom types and positions by maximizing the likelihood of the data
under the forward diffusion process. The model predicts $\mathbf{C}_{\text{pred}}$, $\mathbf{A}_{\text{pred}}$, and
$\mathbf{B}_{\text{pred}}$  from noisy inputs $(\mathbf{A}_\tau, \mathbf{B}_\tau, \mathbf{C}_\tau)$ at a randomly sampled timestep $\tau$.

\subsubsection{Loss Functions}
\label{subsubsec:loss}

Our loss function combines multiple components:

\textbf{Atom Type Loss}: Cross-entropy with label smoothing (smoothing factor 0.02):
\begin{equation}
\mathcal{L}_{\text{type}} = \text{CrossEntropy}(\mathbf{A}_{\text{pred}}, \mathbf{A}_{\text{true}}; \text{smoothing}=0.02)
\end{equation}

\textbf{Bond Type Loss}: Cross-entropy on bonds:
\begin{equation}
\mathcal{L}_{\text{bond}} = \text{CrossEntropy}(\mathbf{B}_{\text{pred}}, \mathbf{B}_{\text{true}}; \text{smoothing}=0.001)
\end{equation}

\textbf{Position Loss}: Mean squared error:
\begin{equation}
\mathcal{L}_{\text{pos}} = \frac{1}{N} \sum_{i=1}^N \|\mathbf{C}_{\text{pred}}^{(i)} - \mathbf{C}_{\text{true}}^{(i)}\|^2
\end{equation}

\textbf{LDDT Loss}: We use the same loss as many other protein-folding models, in order to ensure good bond distances, bond
angles, dihedral angles, and other multi-atom distances.
\begin{equation}
\mathcal{L}_{\text{lddt}} = 1 - \frac{1}{N(N-1)} \sum_{i \neq j} \frac{1}{|\mathcal{T}|} \sum_{\tau \in \mathcal{T}} \sigma\!\left(\tau - \left|\|\mathbf{C}_{\text{pred}}^{(i)} - \mathbf{C}_{\text{pred}}^{(j)}\| - \|\mathbf{C}_{\text{true}}^{(i)} - \mathbf{C}_{\text{true}}^{(j)}\|\right|\right)
\end{equation}
where $\mathcal{T} = \{0.5, 1.0, 2.0, 4.0\}$ Å are distance thresholds and $\sigma(x) = 1/(1+e^{-x})$ is the logistic sigmoid.

The total loss is a simple unweighted combination:
\begin{equation}
\mathcal{L}_{\text{total}} = \mathcal{L}_{\text{type}} + \mathcal{L}_{\text{bond}} + \mathcal{L}_{\text{pos}}+\mathcal{L}_{\text{lddt}}
\end{equation}

\subsubsection{Pretraining Procedure}
\label{subsubsec:training_proc}

\textbf{Density Map Augmentation}: During training, we apply augmentation to density maps:
\begin{itemize}
    \item 50\% dropout: Zero out the density map (unconditioned generation)
    \item 50\% unchanged: Use original density map
\end{itemize}

This augmentation encourages the model to be robust to missing or noisy conditioning information. It also ensures that the
model does not over-index on the existence of protein or ligand maps, as the information contained therein is non-trivial.
Datapoints with no protein map necessarily come from the ligand-only data source, while datapoints with no ligand map come
from the protein-ligand data source, but had an error during ligand map generation, which is strongly correlated with the
presence of certain atoms, such as Boron or Silicon, for which the forcefield generation can fail.

\textbf{Optimization}: We use AdamW optimizer with learning rate $2\cdot10^{-3}$ and $\epsilon = 10^{-6}$. Notably, we use a higher-than-default weight decay of 0.1. Learning rate scheduling:
\begin{itemize}
    \item Linear warmup: 500 steps, start factor $10^{-3}$
    \item Cosine decay: $T_{max}=$100k, $\eta_{min}=10^{-3}$
\end{itemize}
We used a batch size of 128, and an epoch size of 8192, leading to a set of validation metrics every 64 optimizer steps. 

\begin{table}[htbp]
\centering
\caption{Training hyperparameters}
\label{tab:training_params}
\begin{tabular}{lr}
\toprule
Parameter & Value \\
\midrule
Learning rate & $2 \times 10^{-3}$ \\
Optimizer & AdamW ($\epsilon = 10^{-6}$) \\
Weight decay & 0.1 \\
Batch size & 128 \\
Epoch size & 8192 \\
Max optimizer steps & 100,000 \\
Gradient norm clipping & 10 \\
\bottomrule
\end{tabular}
\end{table}

\subsubsection{Trajectory finetuning}
\label{subsubsec:reflow_training}
To aid the model in creating high quality molecules during generation, we applied a fine-tuning pass during which we perform the following procedure:

\begin{itemize}
    \item Sample a datapoint identically to that in pretraining.
    \item Noise that datapoint to a random noise level in the full noise space.
    \item Perform a 100-step diffusion process with that starting noised datapoint, recording each intermediate state as a separate datapoint.
    \item Perform a matching algorithm from the ground truth to the final prediction, and use that permutation to re-order the ground truth.
    \item Emit each pair of (input, permuted ground truth) as a datapoint for finetuning.
\end{itemize}

The matching step is critical, because although the individual atom slots have fixed identity, as they were noised from a given input, often a pretrained model will choose to permute several slots early in the trajectory, and not recomputing the permutation leads to a misalignment between the training-time prediction and the ground truth.

For finetuning, we use a learning rate of $2\cdot 10 ^{-6}.$ Other hyperparameters are the same as pretraining.

\subsection{Data Pipeline}
\label{subsec:data}

\subsubsection{Ligand-Only Datasets}
\label{subsubsec:ligand_only}

\textbf{ZINC Database}: We use the ZINC22 database~\citep{tingle2023zinc22}, which contains over 37 billion ready-to-make commercially
available compounds. We have downloaded approximately 15 billion compounds for training, filtered to those with
3--50 heavy atoms. For each compound, we generate a 3D conformer using RDKit, apply MMFF94 force field
initialization, and compute a density map from the resulting forces and RDKit-derived properties.

ZINC provides diverse molecular structures without protein context, enabling the model to learn general molecular
geometry and chemistry.

\subsubsection{Ligand-Protein Datasets}
\label{subsubsec:ligand_protein}

\textbf{SAIR Dataset}: The Structurally-Augmented IC50 Repository~\citep{lemos2025sair} contains
protein-ligand complexes, predominantly co-folded (synthetic) structures rather than experimental ones.
We process SAIR by:
\begin{itemize}
    \item Extracting binding pockets around ligands (radius $8\sqrt{3}$ Å)
    \item Computing protein properties (charge, hydrophobicity, H-bond donors/acceptors, aromaticity, VdW radii)
    \item Generating density maps for both protein and ligand as above
\end{itemize}

\subsubsection{Density Map Generation}
\label{subsubsec:density_gen}

Density maps are generated with a grid resolution of 0.5 Å over a fixed cube size of $32 \times 32 \times 32$ voxels, or a cube with side length 16 \Angstrom{} centered at the molecule's centroid. We used 6 channels encoding charge, hydrophobicity, H-bond donor, H-bond acceptor, aromatic interactions,
    and van der Waals potential.

For ligand density maps, an MMFF94 forcefield was generated using RDKit, and the parameters of that forcefield were used to determine the parameters for the density map generation below. Hydrophobicity channel parameters were determined by using RDKit's Crippen contributions per-atom. Aromaticity was determined from automatic recognition by RDKit, and H-bond acceptor and donors were detected with SMARTS patterns.

For protein density maps, each protein heavy atom had a hardcoded list of the parameters, with no adjustments made for their environment.

The density map generator places atomic contributions onto the grid using Gaussian kernels for the hydrophobicity, $\pi$-stacking, and H-bond channels, with channel-specific kernel widths for the different interaction types. The electrostatic and van der Waals potentials were computed using their standard analytic formulas. For the van der Waals channel, a post-processing transform of a scaled tanh was applied for numerical stability.

\subsection{Data Augmentation}
\label{subsec:augmentation}

\subsubsection{Rotational Augmentation}
\label{subsubsec:rotation}
Random 3D rotations are applied to both ligands and proteins, ensuring rotational invariance. Rotations are
generated using QR decomposition of random matrices, ensuring uniform sampling from SO(3). This augmentation
prevents overfitting to specific orientations.

\subsubsection{Fragment augmentation}
\label{subsubsec:fragment}

In order to obtain a model capable of denoising a molecule with a soft constraint of a molecular fragment, we perform the following data augmentation when generating the ligand data on-the-fly during training.
First, we use RDKit's Murcko scaffold implementation to select the canonical scaffold for the target molecule. Then, we sample uniformly the desired fragment size. If that size is smaller than the scaffold, then we remove atoms one at a time at random, restricting to those that would not create disconnected fragments. If the desired size is larger than the scaffold, we add atoms from the molecule one at a time at random, restricting to those that directly connect to atoms already within the fragment. In this way, we obtain an augmented set of fragments that are uniformly size-distributed, and heavily biased towards scaffold-like fragments. This fragment is then used to create the ligand density map in 50\% of ligand-conditioned cases. Note that the direct inputs and outputs of the model are unchanged; only the conditioning itself is affected.

\begin{table}[htbp]
\centering
\caption{Dataset statistics}
\label{tab:dataset_stats}
\begin{tabular}{lrl}
\toprule
Dataset & Size & Notes \\
\midrule
ZINC & $\sim$15B (of 37B) & Ligand-only, 3D conformers \\
SAIR & $\sim$8M & Protein-ligand complexes \\
\bottomrule
\end{tabular}
\end{table}

\section{Experiments}
\label{sec:experiments}

\subsection{Training Setup}
\label{subsec:setup}

Training was performed on an A100 GPU. We used PyTorch 2.7.1 with CUDA 12.6 support and PyTorch Lightning 2.5.0. Pretraining ran for approximately a week over 100k optimizer steps, which corresponds to ~12M training datapoints. Finetuning ran for approximately 24 hours, over ~1M datapoints.

\subsection{Training Validation}
\label{subsec:validation}

We validate the model across multiple conditioning modes to assess its ability to use structural information:

\begin{itemize}
    \item \textbf{Fully conditioned (P+L)}: Both protein and ligand density maps provided
    \item \textbf{Protein and fragment conditioned (P+F)}: Both protein density maps and a ligand density map from a fragment provided
    \item \textbf{Protein-only (P)}: Only protein density map provided
    \item \textbf{Full ligand conditioning (L)}: Only ligand density map provided
    \item \textbf{Fragment conditioning only (F)}: Only a ligand density map from a fragment is provided
    \item \textbf{Unconditioned (U)}: No density maps provided
\end{itemize}
Note that LDDT here is a sigmoid-normalized fraction of atom-pairs that correctly lie within [0.5, 1, 2, 4] \Angstrom of the ground truth; a surrogate to the loss used in training. Single-fragment is the fraction of molecules that are fully-connected, which proves to be a surprising challenge for variable-atom count models. Atom validity is the proportion of atoms that have correct valence structure. Note that Sesame splits out different atom types according to their hydrogen count, so this metric is stricter than the same metric for many other similar models. Molecule validity is the fraction of generated molecules that satisfy both of the above, for all the constituent atoms.
Table \ref{tab:val_metrics_pretrained} shows the results for pretraining, and table \ref{tab:val_metrics_finetuned} shows the results for finetuning. Notably, trajectory-level metrics improve with finetuning, while single-step metrics stay roughly equivalent or degrade slightly.

\begin{table}[htbp]
\centering
\caption{Validation metrics by conditioning mode (pretrained)}
\label{tab:val_metrics_pretrained}
\begin{tabular}{lrrrrrr}
\toprule
 & P+L & P+F & P & L & F & U \\
\midrule
\multicolumn{7}{c}{Single-step metrics from randomly partially noised inputs} \\
\midrule
Loss & 0.349 & 0.508 & 0.684 & 0.349 & 0.565 & 0.857 \\
Type Accuracy & 0.983 & 0.955 & 0.924 & 0.983 & 0.949 & 0.909 \\
Bond Accuracy & 0.997 & 0.995 & 0.993 & 0.997 & 0.995 & 0.993 \\
True Bond Acc & 0.911 & 0.850 & 0.788 & 0.911 & 0.838 & 0.761 \\
Bond Precision & 0.983 & 0.978 & 0.977 & 0.983 & 0.977 & 0.982 \\
Bond Recall & 0.915 & 0.859 & 0.803 & 0.915 & 0.848 & 0.776 \\
Position MSE & 0.022 & 0.074 & 0.135 & 0.022 & 0.105 & 0.241 \\
LDDT & 0.982 & 0.964 & 0.946 & 0.982 & 0.958 & 0.929 \\
\midrule
\multicolumn{7}{c}{20-step reverse diffusion validation metrics} \\
\midrule
Single-fragment & 0.991 & 0.939 & 0.857 & 0.989 & 0.924 & 0.816 \\
Atom validity & 0.990 & 0.992 & 0.996 & 0.990 & 0.991 & 0.988 \\
Molecule validity & 0.755 & 0.776 & 0.773 & 0.752 & 0.753 & 0.734 \\
\bottomrule
\end{tabular}
\end{table}

\begin{table}[htbp]
\centering
\caption{Validation metrics by conditioning mode (finetuned)}
\label{tab:val_metrics_finetuned}
\begin{tabular}{lrrrrrr}
\toprule
 & P+L & P+F & P & L & F & U \\
\midrule
\multicolumn{7}{c}{Single-step metrics from randomly partially noised inputs} \\
\midrule
Loss & 0.381 & 0.532 & 0.692 & 0.380 & 0.594 & 0.868 \\
Type Accuracy & 0.978 & 0.950 & 0.923 & 0.978 & 0.944 & 0.908 \\
Bond Accuracy & 0.996 & 0.995 & 0.993 & 0.996 & 0.995 & 0.993 \\
True Bond Acc & 0.919 & 0.855 & 0.790 & 0.919 & 0.845 & 0.760 \\
Bond Precision & 0.976 & 0.969 & 0.973 & 0.976 & 0.970 & 0.982 \\
Bond Recall & 0.923 & 0.866 & 0.805 & 0.923 & 0.856 & 0.774 \\
Position MSE & 0.030 & 0.080 & 0.138 & 0.030 & 0.115 & 0.246 \\
LDDT & 0.980 & 0.963 & 0.945 & 0.980 & 0.956 & 0.929 \\
\midrule
\multicolumn{7}{c}{20-step reverse diffusion validation metrics} \\
\midrule
Single-fragment & 0.994 & 0.964 & 0.899 & 0.994 & 0.950 & 0.820 \\
Atom validity & 0.992 & 0.993 & 0.996 & 0.992 & 0.992 & 0.991 \\
Molecule validity & 0.796 & 0.805 & 0.825 & 0.792 & 0.791 & 0.763 \\
\bottomrule
\end{tabular}
\end{table}

The main type of error the model makes is creating spurious atoms. This is most noticeable in the fragment-conditioning regime, but occurs across all conditioning environments. These excess atoms are typically singletons that do not have any bonds, and the issue can likely be solved with more finetuning or cleverer noise schedules. However, we note that with simple post-processing to remove these atoms, the validity metrics for the two most salient regimes for drug discovery -- protein + fragment for lead optimization and protein-only for \textit{de novo} generation -- jump to 92.4\% and 88.7\% respectively on the finetuned model.

Beyond producing valid molecules, fragment conditioning must actually steer generation toward the supplied seed. Across all fragment-conditioned generation (both protein + fragment and fragment-only), 94.8\% of generated molecules contain the seeding fragment as a substructure, confirming that the soft density-map conditioning reliably preserves the intended scaffold rather than discarding it.

\begin{figure}[htbp]
\centering
\includegraphics[width=0.95\textwidth]{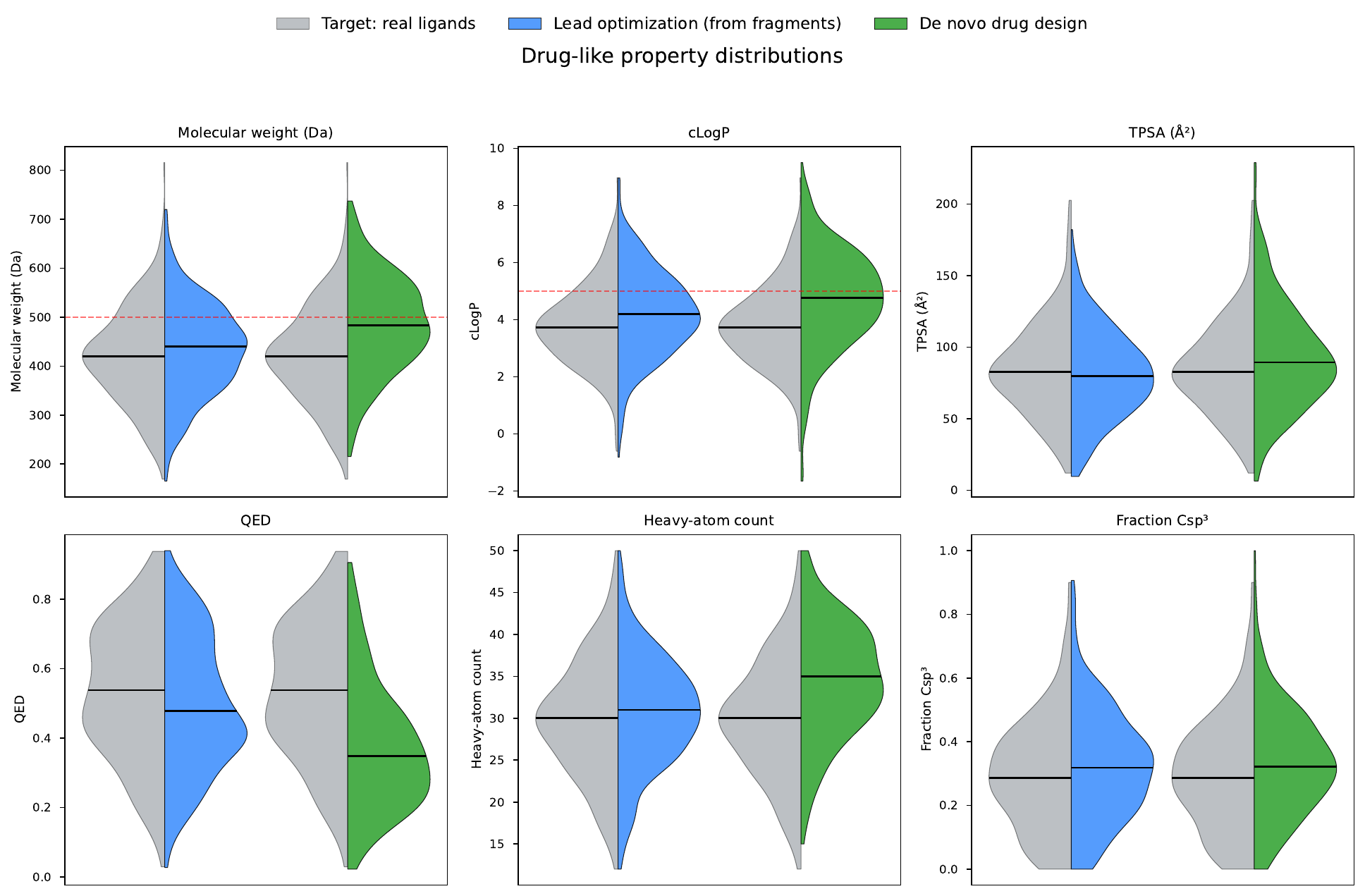}
\caption{\textbf{Sesame produces drug-like molecules.} We compare Sesame's molecule generation to ligands drawn from SandboxAQ's SAIR dataset, which has been widely used to train recent models. Sesame produces molecules with a similar distribution of properties as real drugs; notably, starting with an existing scaffold (blue distributions) performs slightly better than \textit{de novo} generation (green distributions). Red dashed lines indicate Lipinski Rule of 5 thresholds, where applicable.}
\label{fig:property_violins}
\end{figure}

Validity alone, however, does not guarantee that generated molecules resemble viable drug candidates. To assess this, we compare the property distributions of Sesame-generated molecules against ligands drawn from SandboxAQ's SAIR dataset \citep{lemos2025sair}, a large synthetic structural dataset that has been widely used to train recent structure-based models. As shown in Figure~\ref{fig:property_violins}, Sesame produces molecules whose property distributions closely track those of real drug-like ligands and largely respect the Lipinski Rule of 5 thresholds. Notably, seeding generation with an existing scaffold yields slightly more drug-like distributions than \textit{de novo} generation.

\subsection{Sampling}
\label{subsec:sampling}

\begin{figure}[htbp]
\centering
\includegraphics[width=0.95\textwidth]{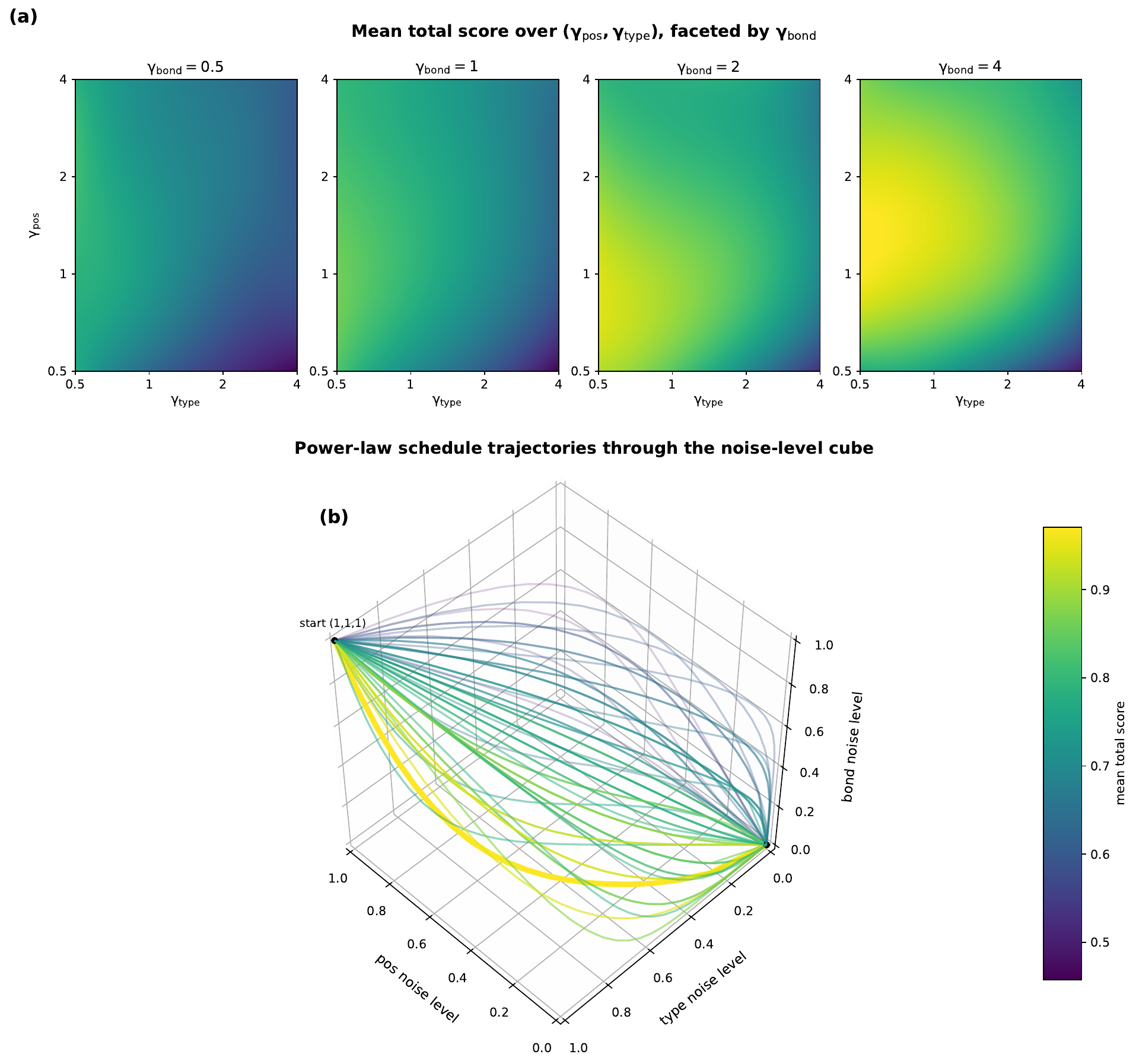}
\caption{Results from the schedule sweep. Top: gradient-smoothed average score from the 64 schedules. Bottom: 3d path of the schedule in normalized noise space. Paths are colored according to their average final score.}
\label{fig:schedule_sweep}
\end{figure}

When sampling with a Sesame model, there are several distinct choices that can be made. The most salient of these is to choose the full 3-dimensional noise schedule, as the model is trained to denoise in a schedule-agnostic manner, but not all paths through noise space are equal. While a full optimization of this path is certainly an option, we chose to adopt a simple strategy: Given that each noise coordinate is normalized to $[0, 1],$ choose a function $\sigma(t) = (t^{\gamma_\text{position}}, t^{\gamma_\text{atom}}, t^{\gamma_\text{bond}}).$ We can then sweep through a small set of choices for each of the $\gamma$ by running a 20-step denoising trajectory with a linear $t$ from 1 to 0. For our sweep, we chose 4 values of $\gamma$ (0.5, 1, 2, 4), to look at 64 schedules with 2048 randomly sampled initial states, protein conditioning, and no ligand conditioning. We scored the generated molecules with a custom score that penalizes the above metrics in a smooth manner. Results of this sweep are in Figure~\ref{fig:schedule_sweep}. As expected, the ideal $\gamma_\text{position}$ was larger than $\gamma_\text{atom}$. This essentially states that fixing positions first is preferable to fixing atom types, and then trying to match them to the correct point in space. Surprisingly, a large $\gamma$ for bonds was strongly favored, one that implies that bonds are fixed before positions become well-determined. Our leading hypothesis is that this is due to two main factors: first, the pairwise matrix of bonds has a magnitude of order more micro-decisions to make, but most of these are relatively trivial. As a result, fixing most bonds to be the non-bonding class is easier to set early and correct at low noise levels, rather than one-shotting the prediction near the end of the diffusion process. The second factor that we believe to be important is that the bond matrix at low noise provides a strong bias for the general structure of the molecule, and make both position and type denoising easier.

\section{Discussion}
\label{sec:discussion}

\paragraph{Human-in-the-loop lead optimization.}
A central design goal of Sesame is to make lead optimization a first-class mode of use rather than a special case of
\textit{de novo} generation. The most common real-world drug-discovery task is not designing a molecule from nothing,
but improving a known hit. Because a partial ligand enters the model as the same continuous density field used for the
protein pocket (Section~\ref{subsubsec:fragment}), a chemist can prune a hit to the substructure worth keeping and hand
Sesame that fragment as a soft prior; the model then grows the remaining chemistry to fit both the fragment and the
pocket. Unlike methods that fix a scaffold as discrete atoms, the field-based prior can be partial and approximate, and
the surrounding atoms, bonds, and geometry are all free to adapt around it.

This dual capability is reflected in our validation. The protein-plus-fragment (P+F) regime is among the strongest
across the trajectory-level metrics (Table~\ref{tab:val_metrics_finetuned}), and after simple post-processing the two
regimes we consider most salient for drug discovery -- protein-plus-fragment for lead optimization and protein-only for
\textit{de novo} generation -- reach 92.4\% and 88.7\% molecule validity respectively. We view this combination, exposed
through a single conditioning interface, as the practical core of the system: human structural insight and generative
chemistry are composed rather than traded off.

\section{Conclusion}
\label{sec:conclusion}

We presented Sesame, a structure-aware molecular generation system that uses density map conditioning and hybrid diffusion.
Our key innovation is an attention-based density map conditioning architecture that adaptively extracts structural
information to guide generation. The hybrid diffusion process handles both discrete atom types and continuous coordinates
within a unified framework.

Experiments demonstrate that the model effectively uses conditioning information, with performance improving when both
protein and ligand density maps are provided. Multi-source training enables learning from both general molecular
chemistry and specific binding interactions.

Critically, because the pocket and a partial ligand are presented to the model through the same density-map interface,
Sesame supports both \textit{de novo} design and fragment-conditioned lead optimization within one framework. Our
validation across protein-only and protein-plus-fragment regimes shows the model makes productive use of a supplied
scaffold, pointing toward a workflow in which a medicinal chemist's structural insight directly seeds generation and
the model elaborates it into pocket-compatible molecules.

% Bibliography
\bibliographystyle{plainnat}
\bibliography{references}

\newpage
\appendix

\section{Density Map conditioning Operations}
\label{app:density_map}

Given single representation $\single \in \mathbb{R}^{B \times N \times d_s}$ where $B$ is batch size, $N$ is number of atoms,
and $d_s=384$ is the single dimension, we compute:
\begin{align}
\mathbf{Q}_{\text{sample}} &= \text{Linear}(\single) \in \mathbb{R}^{B \times N \times H \times d_h} \\
\mathbf{K}_{\text{sample}} &= \mathbf{K}_{\text{learned}} \in \mathbb{R}^{H \times O \times d_h} \\
\mathbf{V}_{\text{sample}} &= \text{Linear}(\single) \in \mathbb{R}^{B \times N \times H \times 3}
\end{align}
where $H=4$ is the number of attention heads, $O = 1024/H = 256$ is the number of sampling points per head (for a total
of $1024$ sampling points), and $d_h = d_s/H = 96$ is the dimension per head. The attention mechanism computes:
\begin{equation}
\mathbf{A} = \softmax\left(\frac{\mathbf{K}_{\text{sample}} \mathbf{Q}_{\text{sample}}^T}{\sqrt{d_h}}\right)
\in \mathbb{R}^{B \times H \times O \times N}
\end{equation}
and the sampling points are computed as:
\begin{equation}
\mathbf{P}_{\text{sample}} = \tanh\left(\text{reshape}(\mathbf{A} \mathbf{V}_{\text{sample}})\right) \in \mathbb{R}^{B \times O \cdot H \times 3}
\end{equation}
where the reshape combines the head and sampling point dimensions. The tanh ensures the range in $[-1, 1]$ for grid sampling.

\textbf{Grid Sampling}: The sampling points are used to extract features from the density map via trilinear interpolation:
\begin{equation}
\mathbf{F}_{\text{density}} = \text{GridSample}(\density_{\text{protein}}, \mathbf{P}_{\text{sample}})
\end{equation}
where $\density_{\text{protein}} \in \mathbb{R}^{d_d \times 12 \times 12 \times 12}$ is the post-3D convnet density map.

\textbf{Cross-Attention}: The $d_d$ density features are concatenated with sampling point coordinates
and used in cross-attention to update single and pair representations:
\begin{equation}
\mathbf{F}_{\text{combined}} = [\mathbf{F}_{\text{density}}, \mathbf{P}_{\text{sample}}] \in \mathbb{R}^{B \times 1024 \times (d_d+3)}
\end{equation}
where $d_d=384$ is the processed density map channel dimension. The cross-attention computes:
\begin{align}
\mathbf{Q}_{\text{single}} &= \text{Linear}(\single) \in \mathbb{R}^{B \times N \times H \times d_h} \\
\mathbf{Q}_{\text{pair}} &= \text{Linear}(\pair) \in \mathbb{R}^{B \times N \times N \times H \times d_h} \\
\mathbf{K}_{\text{density}} &= \text{Linear}(\mathbf{F}_{\text{combined}}) \in \mathbb{R}^{B \times 1024 \times H \times d_h} \\
\mathbf{V}_{\text{single}} &= \text{Linear}(\mathbf{F}_{\text{combined}}) \in \mathbb{R}^{B \times 1024 \times d_s} \\
\mathbf{V}_{\text{pair}} &= \text{Linear}(\mathbf{F}_{\text{combined}}) \in \mathbb{R}^{B \times 1024 \times d_p}
\end{align}
where $\pair \in \mathbb{R}^{B \times N \times N \times d_p}$ is the pair representation with $d_p=128$. The attention outputs are:
\begin{align}
\single_{\text{conditioned}} &= \text{reshape}(\softmax(\mathbf{Q}_{\text{single}} \mathbf{K}_{\text{density}}^T / \sqrt{d_h}) \mathbf{V}_{\text{single}}) \in \mathbb{R}^{B \times N \times d_s} \\
\pair_{\text{conditioned}} &= \text{reshape}(\softmax(\mathbf{Q}_{\text{pair}} \mathbf{K}_{\text{density}}^T / \sqrt{d_h}) \mathbf{V}_{\text{pair}}) \in \mathbb{R}^{B \times N \times N \times d_p}
\end{align}
The conditioning is applied to the single and pair representations, respectively, through gated residual connections.

\end{document}